\documentclass[preprint,number,sort&compress,12pt]{elsarticle}
\usepackage{amsmath,amssymb}
\usepackage{booktabs,longtable,array,calc}
\journal{Applied Energy}
\providecommand{\tightlist}{\setlength{\itemsep}{0pt}\setlength{\parskip}{0pt}}
\begin{document}
\begin{frontmatter}
\title{From fleet trajectories to single-cell promises: hierarchically weighted conformal prediction for battery SOH monitoring}
\author[a]{Xiangyu Liu}
\ead{tl_lucario@163.com}
\address[a]{Independent Researcher, Beijing, China}
\begin{abstract}
Interval predictions from conformal prediction are usually read as ``90\% coverage'', but on nested battery data that number promises the fleet average, not the individual cell. We measure the gap on 124 public LFP cells under a preregistered three-tier reporting protocol (fleet, batch, cell): pooled coverage looks healthy (0.875--0.906 across splits) while the worst cell is covered 0.1--2\% of the time, with exact binomial tests against the nominal level. We then repair part of the gap with a hierarchically weighted conformal engine whose weights consume only side information frozen at prediction time --- signature features, batch labels, cycle indices --- and never touch residuals. The preregistered repair criterion passes: 10th-percentile cell coverage moves 0.750\(\rightarrow\)0.830 (53\% of the gap), the worst cell 0.642\(\rightarrow\)0.694, with zero deferrals and a flat sensitivity spectrum. What the engine does not repair is localized rather than left open: under protocol drift the frozen weights harm monotonically on the SOH scale (the absolute scale is reported in full), a structural dead end we report instead of tuning away; and repair capacity is bounded by calibration-pool battery diversity, measured on both datasets. On a cross-chemistry external set the calibrated intervals collapse (0.286), while the within-domain skeleton replicates (pooled 0.933, worst 0.686). The deliverable is a reporting protocol --- three tiers of promise anchored to measured regimes, worst-cell statistics, deferral share --- that makes per-cell interval claims auditable before deployment.
\end{abstract}
\end{frontmatter}
\subsection{Introduction}\label{introduction}

Grid-scale storage and electric-vehicle fleets have moved battery aging out of the laboratory and into operations. A state-of-health (SOH) number now sits inside decisions that carry price tags: warranty thresholds, capacity-fade alarms, module replacement scheduling, second-life triage. When a warranty reserve is priced from an SOH interval, the gap between fleet-grade and cell-grade coverage is a cash number, not a statistical nicety. None of these decisions consumes the point estimate alone. The operator's question is how wrong the number can be --- an interval with a stated error rate, attached to the specific cell under watch.

Conformal prediction sells exactly that commodity without distributional assumptions: wrap any point predictor, calibrate on held-out residuals, emit intervals with a finite-sample coverage guarantee at a chosen level \cite{vovk2005,papadopoulos2002,tibshirani2019}. The deployment story is nearly frictionless --- no retraining, no likelihood, one scalar \(\hat q\) for the whole fleet. The fine print is the guarantee's object. The nominal statement is about a draw from the calibration pool. A 90\% guarantee at fleet level is compatible with one cell being wrong on most of its cycles. Maintenance decisions are cell-level. The headline number can be healthy while the decision-relevant quantity is broken.

Battery calibration data are not an exchangeable pool of independent draws. Cycles nest inside cells; cells nest inside manufacturing and fast-charging-protocol batches; aging and protocol changes both shift the residual distribution over the fleet's life. Within-cell correlation and batch structure inflate the effective sample size behind the pooled quantile, which then tracks the residual scale of typical cells and under-covers cells whose own scale is larger \cite{barber2023,oikarinen2024}. This is not an implementation bug; it is the assumption behind the guarantee meeting the shape of the data. Two independent lines flag it as a live evaluation problem rather than a technicality: same-cell leakage --- cycles of one cell straddling train and test --- measurably inflates validation performance in the battery literature \cite{barua2026}, and Gaussian-process posterior intervals inherit their \(\sigma\) from an optimizer whose failure is invisible in a coverage table (we measure this in §8.4).

We measure the gap, repair part of it, and map the boundary. On the public 124-cell LFP dataset of Severson et al.~\cite{severson2019} --- rebuilt from the original sources, including the documented re-run of five cells --- we run standard split conformal and read coverage at three tiers: fleet, batch, cell (Fig. 1). The exhibit is stark and was preregistered as a two-panel gate. Under random splits, the regime where pooled coverage should hold, the fleet reads 0.906 against a 0.90 target, while 28\% of cells fall below 0.85 and the median worst cell is covered 0.642 of its cycles (exact binomial, Bonferroni over cells, \(p = 4\times10^{-55}\)). On the official split the worst cell is covered 1.4\% of the time: a ``90\%'' interval that is wrong 98.6\% of the time for that cell. Under a protocol-drift split (train on 2017, test on 2018) the fleet number itself collapses to 0.799, and a permutation control attributes the deficit to exchangeability failure rather than sampling noise (recomposed splits cover 0.850--0.940, \(p < 0.001\)).

We then repair what a side-information front-end can repair. The engine re-weights the calibration pool along the hierarchy --- battery similarity, batch, trajectory position, per-cell balance cap --- with every input frozen at prediction time and never touching residuals, the regime under which the weighted-exchangeability bound of Barber et al.~\cite{barber2023} applies. One of three preregistered repair criteria passes: 10th-percentile cell coverage moves from 0.750 to 0.830, closing 53\% of the gap to nominal; the batch tier recovers to 0.884; the fraction of cells below 0.85 halves; deferrals stay at zero and the sensitivity spectrum is flat across the whole tuning range. The two failures are localized rather than open. Under protocol drift the engine harms monotonically on the SOH scale (§8.3 reports both scales row by row) --- a frozen-signature dead end, because repairing it would require weights that read residual scales, which is precisely what the freezing protocol forbids. On the official split the repair is capped by calibration-pool representativeness, and the engine's own cap census says so.

The boundary is mapped on an external dataset: eight Oxford NMC-LCO pouch cells from a different laboratory, chemistry, form factor and load profile \cite{birkl2017}. Cross-domain coverage collapses to 0.286, and the engine does not save it (0.376). Within-domain leave-one-out reproduces the collapse skeleton --- fleet 0.933, worst cell 0.686. Two measurements on two datasets, the cap census on the main dataset and the pool ceiling on the external set, land on the same sentence: the repair capacity of hierarchical weighting is bounded by the diversity of batteries in the calibration pool, not by the number of cycles. A fleet that wants repairable per-cell promises should buy pool diversity, not cycle count.

Our contributions:

\begin{enumerate}
\def\labelenumi{\arabic{enumi}.}
\tightlist
\item
  \textbf{A cell-level coverage benchmark on battery degradation data.} A three-tier reporting protocol (fleet / batch / cell) with a worst-cell exact binomial statistic, permutation attribution, and deferral disclosure. The collapse exhibit: pooled coverage 0.875--0.906 under near-exchangeable splits while the worst cell reads 0.014 (official split) and 0.001 (protocol drift).
\item
  \textbf{A hierarchically weighted conformal engine under a freezing protocol.} Weights from side information available at prediction time; guarantee via the weighted-exchangeability bound; a preregistered partial repair (q10 0.750$\rightarrow$0.830) plus two honest failures with located mechanisms (frozen-signature dead end under drift; calibration-pool representativeness ceiling).
\item
  \textbf{A four-arm comparison with ablations.} Mondrian within-batch CP half-repairs --- q10 is cosmetically restored while the worst cell does not move, the falsification behind the reporting protocol. GP posterior intervals are conservative by non-convergence, not by calibration; a Bayesian hierarchical comparator buys its coverage at 2$\times$ width. Feature families divide the labor across a coverage axis and a width axis.
\item
  \textbf{Boundary mapping.} Cross-chemistry transfer collapse (0.286) reported as a measured limitation; within-domain skeleton replication across five simultaneous differences (chemistry, form factor, laboratory, load profile, time grid); the repair ceiling, measured twice, is the battery diversity of the calibration pool.
\end{enumerate}

\begin{figure}
\centering
\includegraphics[width=\linewidth]{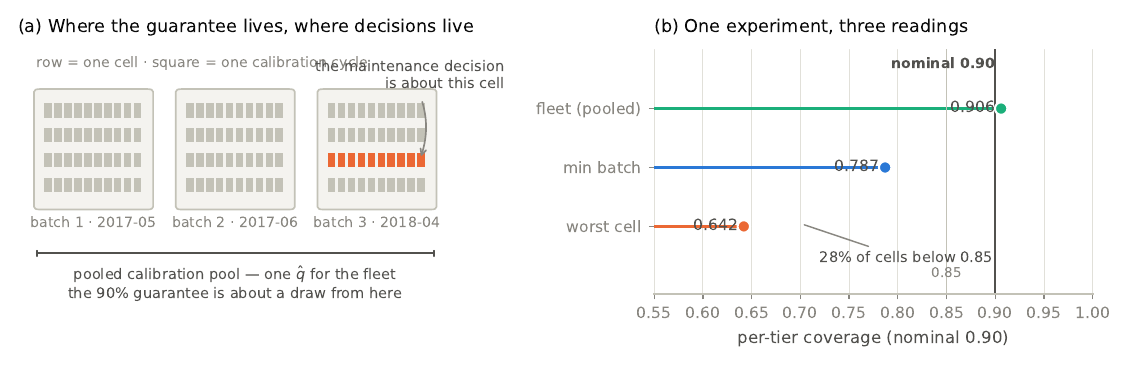}
\caption{The object of the guarantee and the three-tier reading (concept figure; numbers from panel A, SOH scale, Table 1). (a) Calibration cycles nest inside cells; cells nest inside protocol batches. The conformal guarantee attaches to a draw from the pooled cycle set; maintenance decisions attach to one cell. (b) The same experiment read at three tiers: fleet pooled coverage 0.906, minimum batch coverage 0.787, worst single cell 0.642, against nominal 0.90; 28\% of cells fall below 0.85. The headline tier passes while the decision-relevant tier fails.}
\end{figure}

Section 2 positions the work against the conformal, hierarchical-conformal and battery-uncertainty literatures. Sections 3--7 set up the problem, data, engine, comparators and protocol. Section 8 reports; Sections 9--10 discuss and conclude.

\subsection{Related work}\label{related-work}

\subsubsection{Conformal prediction and the object of its guarantee}\label{conformal-prediction-and-the-object-of-its-guarantee}

Conformal prediction produces distribution-free prediction sets with finite-sample validity under exchangeability \cite{vovk2005}. The inductive, split form \cite{papadopoulos2002} made it a deployment primitive: one regression fit, one held-out calibration set, one order statistic --- cheap enough for an embedded monitor. Weighted conformal first appeared for covariate shift, where known density ratios reweight calibration points \cite{tibshirani2019}. Barber et al.~\cite{barber2023} generalized the machinery to arbitrary nonexchangeability: weights in \([0,1]\) buy a coverage bound whose penalty is the normalized mass of misordered calibration points. Their theory is the commodity this paper spends; drift applications are their named future work. We instantiate it where the drift is structural --- nesting, batch protocol --- rather than merely temporal, and we hold the weights to a freezing protocol (§5.5). For time series at large, adaptation and rolling procedures are surveyed by Oikarinen et al.~\cite{oikarinen2024}; their targets are pooled or rolling coverage, not conditional coverage of a nested unit. Perfect conditional coverage is impossible without additional structure \cite{vovk2013,lei2014}; we take that as framing. Normalized scores and conformalized quantile regression \cite{romano2019} address the conditional-variance dimension of this problem; the dimension addressed here --- pool representativeness across nested units --- is orthogonal to that remedy and not substituted by it. The deliverable is a measured gap, a partial repair inside the guarantee, and a reporting protocol --- not an attack on the impossibility.

\subsubsection{Hierarchical and group-aware conformal prediction}\label{hierarchical-and-group-aware-conformal-prediction}

Three lines carry hierarchy into conformal prediction; none operates at cell granularity on degradation data. Dunn et al.~\cite{dunn2023} build distribution-free prediction sets for two-layer hierarchical models by CDF pooling and subsampling --- observations nested in groups, the right nesting shape, without degradation dynamics, drift weights, or a cell-conditional reporting object. Principato et al.~\cite{principato2024} enforce hierarchy as forecast-reconciliation coherence, where aggregates are linear combinations of components --- a different object than unit nesting, despite the shared word. Closest in spirit, the HCP line \cite{lee2026,mallick2026} builds distribution-free inference for grouped data, including unseen groups, with donor groups and initial observations restoring symmetry. Their groups are static and cross-sectional --- hospitals, surveys; ours sit on degradation trajectories, where the commitment object is per-cell conditional coverage and the repair must survive protocol drift, which donor symmetry does not address. The battery-domain analogue of an ``initial sample'' is a new cell's early cycles, which makes the donor route a natural complement to frozen weights once warm-up data exist \cite{attia2020}; we treat it as such, not as a competitor (§9). The same group-aware pattern is being productized in other applied domains --- length-of-stay prediction across hospitals \cite{shahbazi2026} --- none of it in energy storage, and none reporting a nested-unit conditional-coverage failure exhibit. Our position in this narrowing space is the cell-conditional commitment object on degradation trajectories; we report the boundary with the same instrument that reports the results.

\subsubsection{Battery health, uncertainty, and evaluation}\label{battery-health-uncertainty-and-evaluation}

The dataset anchoring this paper comes with its own uncertainty tradition. Severson et al.~\cite{severson2019} built cycle-life prediction from early-cycle features on these cells; the companion tooling reports Gaussian-process posterior intervals, and we measure what that \(\sigma\) is worth under non-convergence and under drift (§8.4). Attia et al.~\cite{attia2020} closed the loop from early-cycle prediction to charging-protocol optimization --- decisions made on short histories, where interval quality at the cell level is the missing half. Bayesian hierarchical models are the long-standing neighbors: Mishra et al.~\cite{mishra2018} and Zhou and Howey~\cite{zhou2022} pool fleet-level and cell-level parameters for capacity prediction. Our BHM-lite comparator descends from that line by design, so that ``why not Bayesian hierarchical'' is answered by measurement: conservative coverage bought at 2$\times$ width, and a drift-panel ``win'' that decomposes into a 100\% fallback artifact (§8.4). Conformal prediction is entering the battery domain from the transfer side \cite{dasilva2026} and from protocol-auditable knee analysis \cite{pregowska2026} --- concurrent, thinner in structure: no cell-conditional commitment object, no hierarchy, no drift repair. Closest to our boundary measurement, Gül~\cite{gul2026} prices the label cost of valid conformal SOH intervals under operating-condition shift on NASA, Oxford and CALCE cells: a zero-shot collapse of the same shape as ours (0.903$\rightarrow$0.221 against our 0.286 cross-domain --- the same collapse, from an independent effort on a different dataset family), and a target-cell warm-up recovery route that is the third-party precedent for the donor complement of §9. On the evaluation side, Barua et al.~\cite{barua2026} show that same-cell leakage inflates validation and that predictor structure modulates the inflation --- third-party evidence that \emph{whose error is being averaged} is a live, under-specified question in this literature. The three-tier protocol of §7 is our answer to it. Both datasets are public \cite{severson2019,birkl2017}.

\begin{center}\rule{0.5\linewidth}{0.5pt}\end{center}

\subsection{Problem setup and notation}\label{problem-setup-and-notation}

Let cell \(b\) run cycles \(k = 1, 2, \ldots\). Write \(Q_b(k)\) for the discharge capacity measured at cycle \(k\) and \(\mathrm{SOH}_b(k) = Q_b(k)\,/\,\bar Q_b^{(1\text{--}10)}\) for the capacity normalized by the cell's own mean discharge capacity over its first ten cycles. At cycle \(k\), with features \(x_{b,k}\) computed only from cycles \(\le k\), we issue an interval prediction for \(\mathrm{SOH}_b(k+H)\) at level \(1-\alpha\), with \(H=1\) and \(\alpha=0.1\) throughout the main text.

Two scale conventions are carried in parallel: normalized SOH (main) and absolute capacity (robustness). Conclusions are read on the main scale; the robustness scale is reported, not averaged.

The calibration unit is the \textbf{cycle}, and cycles are nested: cycles within a cell are strongly correlated; cells within a batch share a fast-charging protocol and manufacturing vintage. A conformal interval computed over a pooled cycle set inherits the exchangeability of that pool --- which is exactly what aging and protocol drift break.

We therefore report coverage at three granularities:

\begin{itemize}
\tightlist
\item
  \textbf{fleet}: coverage pooled over all test cycles (the standard headline number);
\item
  \textbf{batch}: coverage pooled within a manufacturing/protocol batch;
\item
  \textbf{cell}: coverage within a single cell's test cycles (the granularity at which maintenance decisions are made).
\end{itemize}

The nominal statement ``\(P(y \in \hat C) \ge 1-\alpha\)'' is a fleet-level statement under exchangeability. Perfect cell-conditional coverage is known to be impossible without additional assumptions \cite{vovk2013,lei2014}. Our position is operational rather than adversarial: measure the gap at all three tiers, attribute it, repair what is repairable, and report the rest.

\subsection{Data}\label{data}

\textbf{Severson 124 (main).} The public dataset of Severson et al.~\cite{severson2019}: 124 commercial A123 APR18650M1A LFP cells cycled under varied fast-charging protocols, three batches (2017-05, 2017-06, 2018-04; 41/43/40 cells), 100,501 usable cycles in total. We rebuild the official 124-cell cohort from the original \texttt{.mat} sources, including the documented stitch of the five re-run batch-1 cells (their continuation segments live in the 2017-06 batch file), and verify cell identity and lifetimes against the official split CSVs (124/124 matched, order-preserving). Batch identity and cycling protocol are recorded for every cell. The official split partitions the cohort into train (41), test1 (43) and test2 (40).

\textbf{Features (7, three families).} From each cell's history up to cycle \(k\) (all inputs lagged: index \(k-1\) and earlier):

\begin{itemize}
\tightlist
\item
  \emph{trajectory}: previous normalized capacity \(q_{\mathrm{prev}}\), short-window difference \(\mathrm{slope}_5 = q_{k-1} - q_{k-6}\), long-window slope \(\mathrm{slope}_{20} = (q_{k-1} - q_{k-21})/20\);
\item
  \emph{electrochemical}: previous charge time \(ct_{\mathrm{prev}}\), previous internal resistance \(ir_{\mathrm{prev}}\);
\item
  \emph{environment/position}: previous mean per-cycle temperature \(ta_{\mathrm{prev}}\) (Tavg), cycle index \(k\).
\end{itemize}

These are deliberately simple hand features; the object of study is the interval, not the point forecast, and a plain regressor keeps the coverage question isolated from model capacity. A warm-up of \(W=22\) cycles is enforced; the feature row at the warm-up boundary (cycle $\approx$ 23) serves as the cell's frozen \textbf{signature} used by the similarity weights of §5.

\textbf{Oxford (external).} Eight Kokam 740 mAh pouch cells (NMC-LCO), 40 °C, Artemis dynamic discharge profiles, with a characterization measurement every 100 drive cycles (519 characterization points); Birkl~\cite{birkl2017}; Oxford Battery Degradation Dataset 1~\cite{oxford2017}; dataset DOI 10.5287/bodleian:ko2kdmygg. Used only for the external boundary tests of §8.4: different lab, chemistry, form factor, load profile, and time grid.

\subsection{Hierarchically weighted conformal prediction}\label{hierarchically-weighted-conformal-prediction}

\subsubsection{Baseline: split conformal}\label{baseline-split-conformal}

A ridge regression (\(\lambda = 1\)) is fit on a training pool; calibration residuals \(s_i = |y_i - \hat y_i|\) are computed on a disjoint calibration pool; the prediction interval at a test point is \(\hat y \pm \hat q\) with

\[\hat q = \text{the } \lceil (n_{\mathrm{cal}}+1)(1-\alpha) \rceil \text{-th smallest calibration residual}.\]

This is the deployment-default: one \(\hat q\) for the whole fleet, refreshed as data arrives.

\subsubsection{Weight structure: four factors, four axes}\label{weight-structure-four-factors-four-axes}

The engine replaces the flat calibration pool with a weighted one. Each factor governs exactly one axis, and the axes do not mix:

\[w_i \;=\; \underbrace{\exp\!\Big(\tfrac{-(d^2(z_c, z_b) - d^2_{\min})}{2\sigma^2}\Big)}_{\text{cell similarity, feature space}}
\;\cdot\; \underbrace{\delta^{\,\mathbb 1[\,\mathrm{batch}(i) \ne \mathrm{batch}(c)\,]}}_{\text{batch/protocol}}
\;\cdot\; \underbrace{\min\!\big(1,\ \overline T(a)/T_b(a)\big)}_{\text{cell balance cap}}
\;\cdot\; \underbrace{\rho^{\,|k_i - a|}}_{\text{trajectory position}}\]

for a test point on cell \(c\) at cycle \(a\), over calibration cycles \(i\) on cells \(b\) (cell \(c\) is the test cell; cells \(b\) index the calibration pool; \(d^2(z_c, z_b)\) compares frozen signatures). Four notes:

\begin{enumerate}
\def\labelenumi{\arabic{enumi}.}
\tightlist
\item
  \textbf{Similarity is relative to the nearest neighbour}: \(d^2_{\min}\) is the squared signature distance from the test cell to its nearest calibration cell, so the nearest cell always receives full weight. This kills the degenerate regime where every weight collapses to zero while a scale-free ESS target still pretends to be met.
\item
  \textbf{Time decay acts on the cycle axis, not on the admission axis.} It aligns calibration cycles to the test point's \emph{aging stage} (\(|k_i - a|\), known at prediction time, zero leakage). Cross-batch similarity is handled by the batch factor alone; the two axes belong to two factors and are never conflated.
\item
  \textbf{The balance cap} limits any single cell's total mass at anchor \(a\) to that of a median cell (\(T_b(a)\) = cell \(b\)'s total mass at anchor \(a\); \(\overline T(a)\) = median over the calibration pool). Rich cells cannot out-shout the pool; poor cells are not topped up.
\item
  \textbf{Structurally \(w_i \le 1\)} --- the support assumption of the bound below.
\end{enumerate}

The test point carries \(w_{\mathrm{new}} = 1\): the semantics of ``one fresh cycle at the test point's own position''.

\subsubsection{Weighted quantile, guarantee, and deferral}\label{weighted-quantile-guarantee-and-deferral}

\[\hat q \;=\; \inf\Big\{\, t \;:\; \textstyle\sum_i w_i\, \mathbb 1\{s_i \le t\} \;\ge\; (1-\alpha)\big(\textstyle\sum_i w_i + 1\big) \Big\}.\]

If no finite \(t\) satisfies the inequality --- total calibration mass below \(\alpha^{-1} - 1 = 9\) fresh-cycle equivalents --- the engine \textbf{defers}: it emits an infinite interval and says so. The deferral share is reported as a first-class outcome, never folded into coverage. At the pool sizes of this paper the row-level mass never falls below that line; the zero-deferral columns in Tables 2--3 are disclosure of that fact, not a capability claim.

The coverage statement is bound-backed rather than nominal --- the weighted-exchangeability bound (Theorem 2 of Barber et al.~2023):

\[P\big(y \in \hat C\big) \;\ge\; 1 - \alpha \;-\; \textstyle\sum_i \tilde w_i(\text{misordering}),\]

where the penalty is the normalized mass of calibration points that are not exchangeable with the test point. The engine's job is to move that mass where it is cheap (batch proximity, aging-stage alignment) and to report where it cannot.

\textbf{Degradation consistency.} With every factor neutralized (similarity \(\equiv 1\), \(\delta = 1\), \(\rho = 1\), cap off), \(w_i \equiv 1\) and \(\hat q\) reduces to the split-conformal order statistic of §5.1. The baseline is the engine's zero-configuration limit --- verified numerically: the all-neutral run reproduces the baseline arm's readings digit-for-digit on every panel. The method section can therefore state the relationship as an identity, not an analogy.

\subsubsection{The ESS account, at battery level}\label{the-ess-account-at-battery-level}

The single tuning knob is \(\sigma\), and it is not tuned against coverage. We set \(\sigma\) by bisection so that the \textbf{battery-level effective sample size}

\[\mathrm{ESS}_{\mathrm{batt}}(\sigma) \;=\; \frac{\big(\sum_b m_b\big)^2}{\sum_b m_b^2}, \qquad m_b \;=\; \mathrm{sim}_b \cdot \mathrm{grp}_b \cdot \mathrm{cap}_b,\]

hits a target (head configuration: \(\mathrm{ESS}_{\mathrm{batt}} = 5\) effective calibration batteries; \(\rho = 0.999\) fixed as pure smoothing, carrying no ESS burden). Floor and ceiling are flagged, not hidden: at \(\sigma \to 0\) the engine degenerates to the nearest battery (\(\mathrm{ESS}_{\mathrm{batt}} = 1\)); at \(\sigma \to \infty\) it saturates at \(\mathrm{ESS}(\mathrm{grp} \cdot \mathrm{cap})\); targets beyond the ceiling are reported at the achieved value with a cap flag.

Why battery-level rather than row-level: a row-level Kish ESS has a \emph{floor} set by within-cell correlation, \(\mathrm{ESS}_\infty(\rho) = (1+\rho)/(1-\rho)\). In this pool \(\rho = 0.99\) puts the floor at 199 by the closed form, with realized row-level ESS saturating at 319--337 across panels (§8.2) --- every row-level target below that is unreachable and merely labels the nearest-cell regime. We report this floor as a finding: the hierarchy makes Barber-style streaming ESS targets structurally unattainable at row level, and the battery count is the honest currency of evidence.

\subsubsection{Freezing protocol (where the weights may not look)}\label{freezing-protocol-where-the-weights-may-not-look}

Everything the weights consume --- signature features, batch labels, cycle indices --- is frozen at prediction time and available before it. The weights never touch residuals, outcomes, or any coverage reading; the \(\sigma\) bisection reads calibration-weight ESS only. The engine is a side-information front-end, and this is the answer to ``where do the weights come from'': from quantities whose availability at prediction time is auditable, in the spirit of the election-night weighting of Barber et al.~\cite{barber2023}. Note also what the tuning never sees: \(\sigma\) and \(d^2_{\min}\) are computed from signatures and calibration weights only --- the bisection reads calibration-weight ESS, nothing else --- so every weight is fixed before any test outcome exists. Weights predictable from the calibration pool are what the bound requires.

\subsection{Baselines}\label{baselines}

Three comparators, each answering a standing reviewer question in measured form:

\begin{itemize}
\tightlist
\item
  \textbf{Mondrian within-batch CP} (group-aware): the test cell's batch label selects same-batch calibration residuals. If the test batch is absent from the calibration pool, the method falls back to the pooled \(\hat q\); fallback counts are disclosed. This is the strongest natural ``just condition on the group'' competitor.
\item
  \textbf{Gaussian-process posterior intervals} (Severson's native uncertainty quantifier): RBF + white-noise kernel, hyperparameters by marginal-likelihood restarts, train subsampled to 3,000 rows, interval \(\hat y \pm 1.6449\,\sigma_{\mathrm{post}}\). The contrast point: a posterior interval whose validity rests on the model, not on a distribution-free guarantee.
\item
  \textbf{BHM-lite} (empirical-Bayes partial pooling, lightweight descendant of the Bayesian hierarchical models of Mishra et al.~2018): group variances shrunk toward the global variance with prior strength \(k_0 = 50\) virtual cycles, \(s^2_{g,\mathrm{shrunk}} = (n_g s^2_g + k_0 s^2_{\mathrm{glob}})/(n_g + k_0)\); no calibration batch $\rightarrow$ global-scale fallback.
\end{itemize}

None of the three carries a deferral mechanism (\(\infty \equiv 0\) structurally); table footnotes say so, because ``refuse when evidence is short'' is an engine-only capability.

\subsection{Experimental protocol}\label{experimental-protocol}

\textbf{Splits.} Three protocols, each isolating one failure mode:

\begin{itemize}
\tightlist
\item
  \textbf{A --- near-exchangeable}: random cell-level splits, five seeds (42--46), reported as the median; 74 train / 25 calibration / 25 test cells per seed. The regime where pooled coverage \emph{should} hold --- and where cell-level failure persists anyway.
\item
  \textbf{B --- protocol/time drift}: train + calibrate on the 2017 batches (84 cells, split 42/42), test on the 40-cell 2018 batch. The regime the fleet default silently breaks in.
\item
  \textbf{C --- official split} (Severson's published train/test partition): the official train pool of 41 cells is split 20/21 into fit and calibration, and the 83 official test cells are held out; a reproduction anchor, reported in one line.
\end{itemize}

Calibration and test cells are disjoint in every panel and every seed; all splits are cell-level set partitions.

\textbf{Metrics.} Per arm and panel: pooled coverage; minimum batch coverage; the 10th percentile of per-cell coverage (q10, descriptive only); the \textbf{worst cell} with an exact one-sided binomial test against \(H_0: p = 0.90\), Bonferroni-corrected over cells; the fraction of cells below 0.85; deferral share; mean \(\hat q\) (width); achieved ESS. The per-cell binomial tests treat a cell's cycles as independent draws; worst-cell significance is therefore re-examined under a block bootstrap and reported alongside (§8.1). q10 is retired as a gate statistic --- it is a batch-granularity statistic and can be repaired while the worst cell does not move (§8.3 shows this happening); it is reported because reviewers ask for it. The division of labor is deliberate and was registered in advance: the worst-cell binomial is the \emph{gate} --- a collapse detector whose power is ample (\(p = 4\times10^{-55}\) on panel A) --- while q10 is the \emph{repair} criterion, a distribution-shape measure chosen because at \(n = 25\) test cells per seed the worst cell is too noisy to grade a repair.

\textbf{Preregistration discipline.} All gates and head configurations were registered in writing before the formal runs: the E1 gate was re-registered (v2) after a v1 flaw was found --- the v1 criterion stitched two failure modes into one split and leaned on q10; v2 separates the two panels and replaces the gate statistic with the worst-cell binomial. Exploratory development ran on a separate machine whose numbers are discarded; all numbers in this paper come from single formal runs on the provenance machine, with logs archived. For the E3 round, ten falsifiable predictions (P1--P10) were written before execution and reconciled after; three changed shape under the data, and we report the changes rather than the plan (Table 5).

\textbf{Configuration.} \(\alpha = 0.1\), \(H = 1\) (horizon sweep in §8.5), ridge \(\lambda = 1\), warm-up \(W = 22\). Engine head configuration: \(\mathrm{ESS}_{\mathrm{batt}} = 5\), \(\delta = 0.3\), \(\rho = 0.999\), balance cap on. Sensitivity spectra are reported for every knob, in full.

\textbf{External protocol (Oxford).} Two arms. \emph{Cross-domain}: Severson official training pool $\rightarrow$ all eight Oxford cells (different lab, chemistry, form factor, dynamic load, time grid), with the IR-free transferable feature subset. \emph{Within-domain}: leave-one-out over the eight cells (3 train / 4 calibration per fold), full per-cell coverage distribution. With \(n = 8\) cells, percentile statistics are meaningless by construction; we report the full distribution and per-cell binomial tests, and read it as a skeleton replication test, not a benchmark.

\begin{center}\rule{0.5\linewidth}{0.5pt}\end{center}

\subsection{Results}\label{results}

\subsubsection{The collapse exhibit (preregistered gate)}\label{the-collapse-exhibit-preregistered-gate}

Table 1 reports the baseline split-conformal engine under the two preregistered gate panels. Both fired.

\begin{longtable}{@{} >{\raggedright\arraybackslash}p{(\linewidth - 14\tabcolsep) * \real{0.1250}} >{\raggedright\arraybackslash}p{(\linewidth - 14\tabcolsep) * \real{0.1250}} >{\raggedright\arraybackslash}p{(\linewidth - 14\tabcolsep) * \real{0.1250}} >{\raggedright\arraybackslash}p{(\linewidth - 14\tabcolsep) * \real{0.1250}} >{\raggedright\arraybackslash}p{(\linewidth - 14\tabcolsep) * \real{0.1250}} >{\raggedright\arraybackslash}p{(\linewidth - 14\tabcolsep) * \real{0.1250}} >{\raggedright\arraybackslash}p{(\linewidth - 14\tabcolsep) * \real{0.1250}} >{\raggedright\arraybackslash}p{(\linewidth - 14\tabcolsep) * \real{0.1250}}@{}}
\caption{Baseline split conformal, six gate cells (E1). Cell identifiers read batch, then cell within batch.}\\
\toprule\noalign{}
\begin{minipage}[b]{\linewidth}\raggedright
scale
\end{minipage} & \begin{minipage}[b]{\linewidth}\raggedright
panel
\end{minipage} & \begin{minipage}[b]{\linewidth}\raggedright
pooled
\end{minipage} & \begin{minipage}[b]{\linewidth}\raggedright
min batch
\end{minipage} & \begin{minipage}[b]{\linewidth}\raggedright
worst cell
\end{minipage} & \begin{minipage}[b]{\linewidth}\raggedright
\(p_{\mathrm{bonf}}\)
\end{minipage} & \begin{minipage}[b]{\linewidth}\raggedright
frac \(<0.85\)
\end{minipage} & \begin{minipage}[b]{\linewidth}\raggedright
perm. \(p\)
\end{minipage} \\
\midrule\noalign{}
\endhead
\bottomrule\noalign{}
\endlastfoot
SOH & A (random, 5-seed median) & 0.906 & 0.787 & 0.642 & 4.2e-55 & 28\% & --- \\
SOH & C (official split) & 0.875 & 0.790 & \textbf{0.014} (b2c1) & \(<10^{-300}\) & 35\% & --- \\
SOH & B (2017$\rightarrow$2018 drift) & \textbf{0.799} & --- & 0.001 (b3c29) & --- & --- & \textless0.001 \\
abs & A & 0.897 & 0.790 & --- & 6.1e-58 & 28\% & --- \\
abs & C & 0.888 & 0.778 & 0.020 (b2c1) & \(<10^{-300}\) & 27\% & --- \\
abs & B & 0.810 & --- & 0.001 (b3c29) & --- & --- & \textless0.001 \\
\end{longtable}

The worst column and the \(p_{\mathrm{bonf}}\) column are two statistics and can belong to different cells: the worst column reports the minimum per-cell coverage, while \(p_{\mathrm{bonf}}\) reports the deepest binomial deficit, which favors long cells. On panel C the minimum-coverage cell is b2c1 (1.4\% of 148 cycles) and the deepest deficit sits on b1c0 (54.6\% of 1829 cycles, \(p\) underflowing); on panel B the minimum-coverage cell is b3c29 (1 of 835 cycles) and the deepest deficit on b3c28 (10.4\% of 518). Both statistics fail on both panels.

Three readings, in increasing order of discomfort. \textbf{(i)} On the near-exchangeable panel A, the fleet number is healthy --- 0.906 against a 0.90 target --- and this is the number a standard evaluation would report and stop at. \textbf{(ii)} One tier down, 28\% of cells sit below 0.85 coverage and the worst cell is covered 64\% of the time. On the official split, the worst cell (b2c1 --- batch 2, cell 1) is covered on \textbf{1.4\%} of its 148 test cycles: a ``90\%'' interval that is wrong 98.6\% of the time for that specific cell. The statistically deepest deficit sits on a different, longer cell (b1c0, one of the five re-run cells: 54.6\% over 1829 cycles, \(p_{\mathrm{bonf}}\) underflowing below \(10^{-300}\)) --- the coverage minimum and the \(p\)-minimum are different cells, and Table 1 reports both. The per-cell coverage distribution (Fig. 2) makes the shape visible: a healthy-looking bulk with a long, disastrous tail. The tail is not a stitching artifact: only three of the five re-run cells sit in the C test set at all, the worst cell is not one of them, and excluding all re-run cells leaves the worst coverage unchanged at 0.014 and the sub-0.85 fraction at 35\% (q10 0.726 $\rightarrow$ 0.734). \textbf{(iii)} Under protocol drift (panel B), the fleet number itself collapses to 0.799, and the permutation control closes the attribution: randomly re-composed splits of the same size cover 0.850--0.940 (5--95\% band), so the deficit is exchangeability failure, not sampling noise (\(p < 0.001\), both scales).

\begin{figure}
\centering
\includegraphics[width=\linewidth]{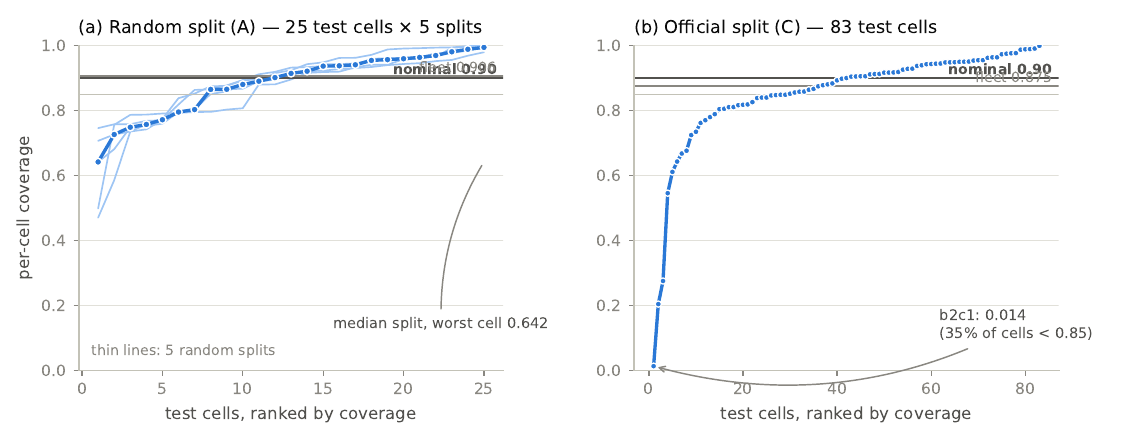}
\caption{Per-cell coverage under baseline split conformal (SOH scale, nominal 90\%). (a) Random split (panel A): five splits (thin) and their pointwise median (bold); the fleet-level number (0.906) sits on nominal while 28\% of cells fall below 0.85 and the median worst cell is 0.642. (b) Official split (panel C): 83 test cells ranked by coverage; the worst cell (b2c1) is covered 1.4\% of the time. A healthy bulk with a disastrous tail is the object the reporting protocol of §7 is built to expose.}
\end{figure}

The failure therefore has two faces --- \emph{conditional} (fleet fine, worst cell broken) under near-exchangeability, and \emph{marginal} (fleet itself broken) under drift --- and both were preregistered as separate gate panels before any formal run. Because a cell's cycle-level coverage events are autocorrelated, the per-cell binomial \(p\)-values are re-examined under a circular block bootstrap (block lengths 5/10/20, \(10^4\) resamples): effective sample sizes shrink --- b2c1: 148 $\rightarrow$ 98, b1c0: 1829 $\rightarrow$ 237 at block length 20 --- yet the panel-B and panel-C worst-cell deficits remain overwhelming (\(|z| \ge 14.7\), \(p \le 6\times10^{-49}\) at block length 20). On panel A the re-examination is honestly weaker: the worst-cell deficit survives Bonferroni in four of five seeds and is marginal in the mildest (worst 0.746, \(p_{\mathrm{bonf}} = 0.07\)) --- one reason the panel-A case rests on the sub-0.85 fraction, a cell-count statistic, and on the full distribution of Fig. 2 rather than on a single cell. The panel-B permutation control resamples cells --- independent physical units --- and is immune to within-cell correlation by construction. The mechanism is the one §3 anticipates: within-cell correlation and batch structure inflate the effective sample size behind the pooled quantile, which then reflects the residual scale of typical cells and systematically under-covers cells whose own residual scale is larger.

\subsubsection{Partial repair under near-exchangeability}\label{partial-repair-under-near-exchangeability}

Table 2 is the preregistered repair headline: baseline vs.~the engine at its registered head configuration (\(\mathrm{ESS}_{\mathrm{batt}} = 5\), \(\delta = 0.3\), \(\rho = 0.999\)), panel A, SOH scale.

\begin{longtable}{@{}lllllll@{}}
\caption{Baseline vs.~engine, panel A (SOH). Registered line: q10 $\geq$ 0.825.}\\
\toprule\noalign{}
arm & pooled & q10 & min batch & worst & frac \(<0.85\) & deferral \\
\midrule\noalign{}
\endhead
\bottomrule\noalign{}
\endlastfoot
baseline & 0.906 & 0.750 & 0.787 & 0.642 & 28\% & 0\% \\
engine & 0.909 & \textbf{0.830} & \textbf{0.884} & 0.694 & \textbf{16\%} & 0\% \\
\end{longtable}

Five-seed ranges behind the medians (q10; worst): baseline 0.738--0.787; 0.470--0.746, engine 0.753--0.858; 0.517--0.838.

The preregistered criterion passes: q10 clears 0.825, closing 53\% of the gap to nominal. The repair lands where the failure was, tier by tier: the batch tier recovers to 0.884, the cell tail moves (worst 0.642$\rightarrow$0.694, sub-0.85 fraction halved), and the fleet tier is untouched (0.906$\rightarrow$0.909). No deferrals were triggered anywhere --- the engine repaired by re-weighting evidence, not by refusing to answer. The seed spread is reported because it is wide: q10 improves on four of five seeds and dips below baseline on one (seed 43: 0.787$\rightarrow$0.753); worst-cell coverage improves on four of five. The medians are the registered readouts, and the spread is the honest noise floor at \(n = 25\) test cells per seed --- one more reason the repair criterion rides on q10 rather than on the noisier worst cell (§7).

The sensitivity spectrum is flat: across \(\mathrm{ESS}_{\mathrm{batt}} \in \{1,3,5,10,20\}\), q10 stays within 0.804--0.830, with no knife-edge and therefore no tuning story for a reviewer to reconstruct (Fig. 3). The head value 5 was registered before the runs; panel C's optimum sits elsewhere (§8.3), and we do not move the head table to chase it. The batch factor is equally flat: \(\delta \in \{0.1, 0.3, 0.5\}\) moves panel-C q10 by at most 0.002 (0.731 / 0.730 / 0.729) and pooled coverage by at most 0.004; \(\rho\) is fixed at 0.999 as pure smoothing by design (§5.4), so the two live knobs are \(\mathrm{ESS}_{\mathrm{batt}}\) and \(\delta\), and both spectra are flat.

\begin{figure}
\centering
\includegraphics[width=\linewidth]{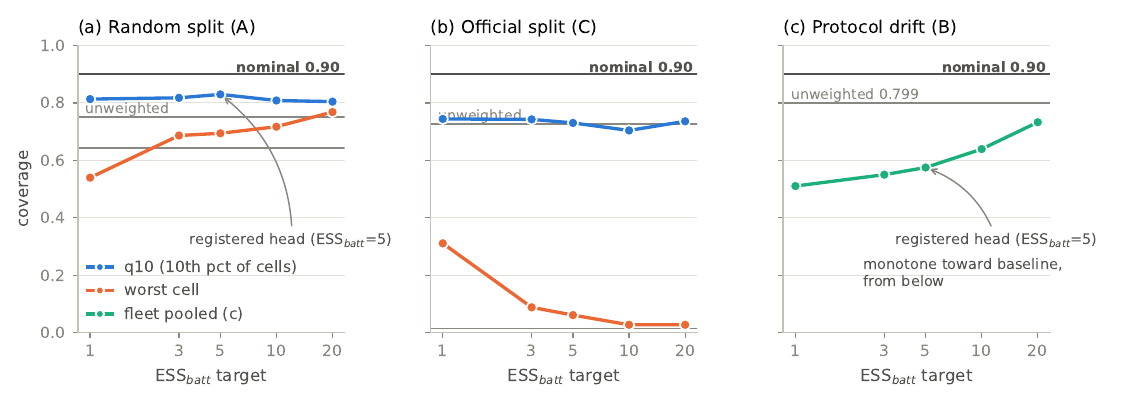}
\caption{Engine sensitivity and the drift dead end (SOH scale). (a, b) q10 and worst-cell coverage vs.~the battery-level ESS target (log axis); gray lines mark the unweighted baseline. Panel A is flat (q10 0.804--0.830) with no knife-edge at the registered head value ESS\_batt = 5; panel C peaks at ESS\_batt = 1 with worst-cell repair 0.014 $\rightarrow$ 0.311. (c) Under protocol drift (panel B) the engine degrades monotonically toward the baseline from below --- the harm is the mechanism, not the setting. Target 20 saturates at achieved ESS $\approx$ 18 (cap flag, §8.3).}
\end{figure}

Two design facts belong on the record because they shape the method. First, the engine's all-neutral configuration reproduces the baseline digit-for-digit on every panel --- the identity of §5.3 holds numerically, so the repair is attributable to the weights, not to incidental code paths. Second, the battery-level ESS knob was itself a redesign after a failed first attempt: a row-level ESS target is structurally floored by within-cell correlation (\(\mathrm{ESS}_\infty(\rho) = (1+\rho)/(1-\rho) = 199\) at \(\rho = 0.99\); the realized row-level ESS of that attempt saturated at 319--337 across panels), so row-level targeting collapsed to the nearest-cell regime and \emph{reversed} the median (q10 0.641, worst 0.433 on panel A). The battery-level redesign keeps the tail repair and restores the median --- the hierarchy's honest currency is batteries, not cycles.

\subsubsection{What does not repair, and why}\label{what-does-not-repair-and-why}

Both remaining preregistered criteria fail, and both failures have a located mechanism rather than an open question.

\textbf{Panel B (protocol drift): a frozen-signature dead end.} The engine \emph{lowers} pooled coverage from 0.799 to 0.575 at the head configuration --- and does so monotonically: ESS\_batt \(1 \to 20\) moves pooled \(0.510 \to 0.550 \to 0.575 \to 0.639 \to 0.733\), approaching the baseline from below, with zero deferrals throughout (Fig. 3). The dose--response says the harm is the mechanism itself, not a bad setting: signature similarity selects \emph{nearest batteries}, and across a protocol boundary the nearest battery by signature is not the nearest by residual scale. Repairing panel B inside the freezing protocol would require weights that read residual scales --- precisely the data-dependent weighting the protocol forbids (§5.5) along with the auditability it buys. We report the curve and the dead end; the escape route is a requalification window, not a weight (§9).

\textbf{Panel C (official split): a calibration-pool ceiling.} q10 reaches 0.730 against a 0.813 line --- a miss. But the cell-tail repair is real and large: at ESS\_batt \(=1\), the worst cell reaches 0.311 versus the baseline's 0.014 --- a 22$\times$ improvement. The ceiling is diagnosable from the machine's own flags: the C calibration pool holds only \textasciitilde20 cells and under-represents the 2018 protocol family, so at ESS\_batt \(=20\) the bisection saturates for 43 of 83 test cells (cap flag), with a maximum achieved effective pool of $\approx$18 batteries. The A and C optima split (A peaks at 5, C at 1); the head table stays at the registered 5. Post-hoc re-pointing would not rescue C anyway --- no configuration passes.

On the absolute scale the same forms hold, slightly weaker (panel A: q10 0.823, 45\% of gap). One honest split is reported row-by-row rather than averaged: on panel B the engine reads 0.575 on SOH but 0.805 on absolute capacity --- a mild improvement over base (0.799) --- so the drift-panel verdict depends on the scale convention, and we say so where the numbers appear.

\subsubsection{Comparators and ablations}\label{comparators-and-ablations}

\begin{longtable}{@{} >{\raggedright\arraybackslash}p{(\linewidth - 12\tabcolsep) * \real{0.1429}} >{\raggedright\arraybackslash}p{(\linewidth - 12\tabcolsep) * \real{0.1429}} >{\raggedright\arraybackslash}p{(\linewidth - 12\tabcolsep) * \real{0.1429}} >{\raggedright\arraybackslash}p{(\linewidth - 12\tabcolsep) * \real{0.1429}} >{\raggedright\arraybackslash}p{(\linewidth - 12\tabcolsep) * \real{0.1429}} >{\raggedright\arraybackslash}p{(\linewidth - 12\tabcolsep) * \real{0.1429}} >{\raggedright\arraybackslash}p{(\linewidth - 12\tabcolsep) * \real{0.1429}}@{}}
\caption{Five arms, panel A (SOH), five-seed medians.}\\
\toprule\noalign{}
\begin{minipage}[b]{\linewidth}\raggedright
arm
\end{minipage} & \begin{minipage}[b]{\linewidth}\raggedright
pooled
\end{minipage} & \begin{minipage}[b]{\linewidth}\raggedright
q10
\end{minipage} & \begin{minipage}[b]{\linewidth}\raggedright
min batch
\end{minipage} & \begin{minipage}[b]{\linewidth}\raggedright
worst
\end{minipage} & \begin{minipage}[b]{\linewidth}\raggedright
width \(\hat q\)
\end{minipage} & \begin{minipage}[b]{\linewidth}\raggedright
fallback
\end{minipage} \\
\midrule\noalign{}
\endhead
\bottomrule\noalign{}
\endlastfoot
baseline & 0.906 & 0.750 & 0.787 & 0.642 & 0.0019 & --- \\
engine (ours) & 0.909 & 0.830 & 0.884 & 0.694 & 0.0021 & 0\% \\
Mondrian (within-batch) & 0.899 & \textbf{0.825} & \textbf{0.869} & \textbf{0.631} & 0.0022 & 0\% (A) \\
GP posterior & 0.965 & 0.913 & 0.932 & 0.786 & 0.0011 & none possible \\
BHM-lite & 0.953 & 0.863 & 0.910 & 0.831 & \textbf{0.0039 (2$\times$)} & 0\% (A) \\
\end{longtable}

Five-seed q10 ranges behind the medians: baseline 0.738--0.787, engine 0.753--0.858, Mondrian 0.796--0.842, GP 0.826--0.996, BHM-lite 0.849--0.994; worst-cell ranges 0.470--0.746 / 0.517--0.838 / 0.570--0.806 / 0.563--0.875 / 0.604--0.971. The medians carry the ranking; the ranges overlap, which is the honest noise floor of a 25-test-cell panel.

\textbf{Mondrian is the half-repair exhibit, sharpened beyond its preregistration.} Conditioning on batch repairs the batch tier (0.869) and \emph{cosmetically} repairs q10 (0.825) --- while the worst cell does not move at all (0.631, \(p_{\mathrm{bonf}}\) 3.2e-59). q10 is a batch-granularity statistic; it can be bought while the worst cell stays broken. This is the falsification behind our reporting protocol: q10 is reported because reviewers ask for it; the worst cell is reported because it is the decision-relevant tier. On panel B, Mondrian's fallback fires for 100\% of test points (the 2018 batch is absent from its calibration pool) and the method degenerates digit-for-digit into the baseline --- group-aware CP structurally has no group to use on an unseen group. Read on panel A alone, the engine's edge over Mondrian is thin --- a few points on most tiers; the engine's reason to exist is the cross-panel account: the tail delta its factors buy (worst 0.694 vs.~0.631), and under drift a measured monotone dose--response (§8.3) where Mondrian silently degenerates into the baseline.

\textbf{GP is conservative by accident, not by calibration.} Its panel-A numbers look superior (0.965 / 0.913 / 0.786) until the machinery is inspected: the kernel hyperparameters ride their upper bound and the optimizer fails to converge, so the posterior $\sigma$ is large by misadventure --- and the worst cell is still below 0.85 (\(p_{\mathrm{bonf}}\) 7.2e-35), with no distribution-free statement attached. On panel B the accident reverses: in the extrapolation regime $\sigma$ explodes (\(\hat q\) = 0.0454, 23$\times$ base width) and pooled 0.814 is width-carried. A posterior interval is only as honest as its $\sigma$, and $\sigma$ here is not under control. The arm runs the toolchain exactly as distributed --- default bounds, default restarts; an ecosystem whose native uncertainty quantifier ships in this state is itself part of the exhibit.

\textbf{BHM-lite pays a width tax, and its drift-panel ``win'' is an artifact.} Panel-A coverage 0.953 is bought at 2$\times$ base width (\(\hat q\) = 0.0039). On panel B it reports 0.994 / 0.957 --- which decomposes into 100\% fallback: no calibration batch exists for the 2018 group, so the shrinkage reverts to a global scale 4$\times$ too wide (0.0082). Coverage without width is half a report; against these comparators, conformal's distribution-free validity \emph{at no width premium} becomes a visible property instead of a slogan.

\textbf{Where the comparators read higher, they pay in width or in an uncontrolled $\sigma$.} On the cell tier, BHM-lite (worst 0.831, q10 0.863) and GP (0.786, 0.913) read above the engine (0.694, 0.830); the prices sit in the same table: 2$\times$ base width for BHM-lite, and for GP a posterior $\sigma$ that rides its optimization bounds rather than converging (as-configured). On this normalized scale the width spread across arms (\(\hat q\) = 0.0011--0.0039, at most $\approx$0.4\% of reference capacity) is operationally small at \(H = 1\), and we do not claim the highest cell-tier reading. The engine's claim is a different one: an auditable distribution-free bound at no width premium, with every width disclosed so the reader can run the cost accounting.

\textbf{Ablations relocate the cause.} sim-only (similarity on; batch, time, balance off) reaches q10 0.823 with no median reversal --- the reversal that killed the row-level variant does not reproduce under a battery-level pool constraint. The cause was pool narrowness, not similarity weighting. bal-only (balance cap alone) sits at baseline on panel A (0.762) while improving panel-B pooled coverage by +0.037 (0.836): pure cell balancing is cheap and mildly drift-tolerant; similarity is what repairs the cell tail; the full engine needs both. Plainly: on panel A the similarity factor does almost all of the repair (sim-only 0.823 vs.~full 0.830); the batch factor's visible work is on the batch tier and on panels B and C, not on the A-panel cell tier.

\textbf{Feature families divide the labor across two axes.} Removing the trajectory family (q\_prev, slopes) leaves q10 apparently \emph{better} (0.870) while width explodes 29$\times$ (\(\hat q\) = 0.0544) and the worst cell collapses (0.449; panel C worst 0.000) --- the trajectory family carries the \emph{width} axis: without it, point prediction degrades, the interval inflates, and coverage survives as an artifact of width. Removing the environment/position family (temperature, cycle index) is the sharpest coverage hit on panel C (q10 0.726$\rightarrow$0.543): the cycle index carries batch signature, echoing the drift mechanism of §8.1. The electrochemical family is mild on both axes. Fig. 4 plots the two axes side by side --- ``which family hurts most'' has no single answer until coverage and width are read together.

\begin{figure}
\centering
\includegraphics[width=\linewidth]{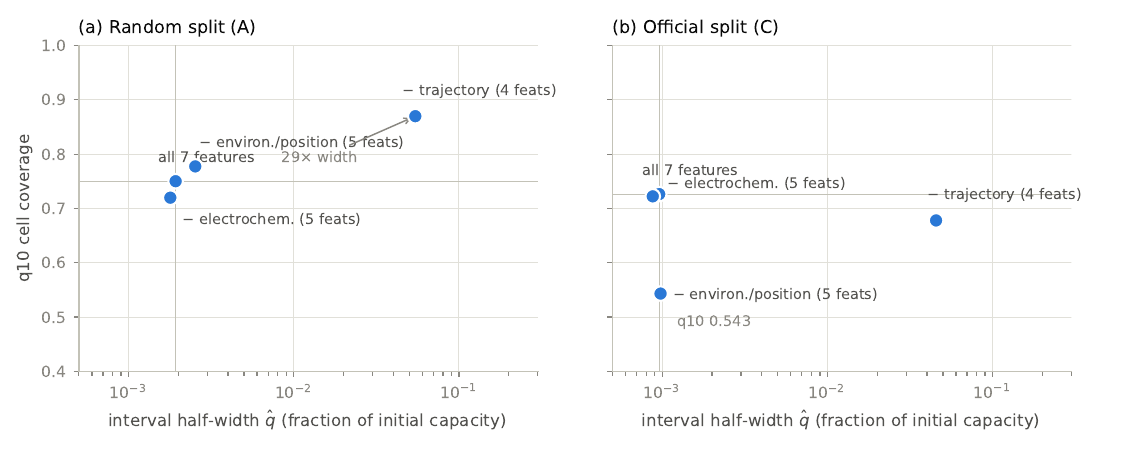}
\caption{Feature families divide the labor across two axes (baseline arm, SOH; interval half-width on log scale). (a) Panel A: removing the trajectory family leaves q10 apparently better (0.870) at 29$\times$ width --- coverage surviving as an artifact of width. (b) Panel C: removing the environment/position family collapses q10 (0.726 $\rightarrow$ 0.543) at unchanged width --- the cycle index carries batch signature, echoing §8.1. The electrochemical family is mild on both axes: its point sits essentially on the full-feature anchor.}
\end{figure}

\subsubsection{Horizon}\label{horizon}

Widening the forecast horizon does not degrade coverage monotonically. The worst cell reads 0.642 (H=1) $\rightarrow$ 0.657 (H=10) $\rightarrow$ 0.539 (H=50): a slight \emph{improvement} at H=10, real degradation at H=50. Width, by contrast, grows monotonically: 0.0019 $\rightarrow$ 0.0034 $\rightarrow$ 0.0118 (6$\times$). We report both axes and drop the single-axis ``farther is worse'' reading; an earlier monotone reading from a development machine did not replicate on the formal runs --- which is §7's discipline doing its job before a reviewer could do it for us.

\subsubsection{External boundary: Oxford}\label{external-boundary-oxford}

\begin{longtable}{@{} >{\raggedright\arraybackslash}p{(\linewidth - 6\tabcolsep) * \real{0.2500}} >{\raggedright\arraybackslash}p{(\linewidth - 6\tabcolsep) * \real{0.2500}} >{\raggedright\arraybackslash}p{(\linewidth - 6\tabcolsep) * \real{0.2500}} >{\raggedright\arraybackslash}p{(\linewidth - 6\tabcolsep) * \real{0.2500}}@{}}
\caption{Oxford, four arms (SOH, IR-free feature subset, n = 8 cells; full distributions, no percentiles).}\\
\toprule\noalign{}
\begin{minipage}[b]{\linewidth}\raggedright
arm
\end{minipage} & \begin{minipage}[b]{\linewidth}\raggedright
pooled
\end{minipage} & \begin{minipage}[b]{\linewidth}\raggedright
worst
\end{minipage} & \begin{minipage}[b]{\linewidth}\raggedright
reading
\end{minipage} \\
\midrule\noalign{}
\endhead
\bottomrule\noalign{}
\endlastfoot
cross-domain plain (Severson pool $\rightarrow$ Oxford) & \textbf{0.286} & 0.137 (oxc2) & \(p_{\mathrm{bonf}}\) 4.5e-36; 100\% of cells below 0.85 \\
cross-domain engine & 0.376 & 0.216 & +0.09; not saved \\
within-domain plain (LOO $\times$8) & 0.933 & \textbf{0.686} (oxc2) & oxc2 binomial \(p\) 2.3e-05; Bonferroni 1.9e-04 \\
within-domain engine & 0.933 & 0.686 & identical to plain; ESS\_batt ceiling 3.73 (cap flag) \\
\end{longtable}

\textbf{The cross-domain arm is the limitation section made empirical.} A ``90\%'' interval calibrated on the Severson fleet is wrong 71\% of the time on Oxford --- feature distributions relocate wholesale (slope magnitudes differ by an order of magnitude; a Severson-calibrated \(\hat q\) of 0.0008 is meaningless against another lab's physics). The engine adds +0.09, which changes no decision. This is the third tier of the signature line in §9: across a laboratory boundary, no interval promise made here survives, and we decline to dress the number up.

\textbf{The within-domain arm is the generalization claim.} Pooled coverage 0.933 with worst cell 0.686 (Oxford cell 2, oxc2): the E1 skeleton --- fleet healthy, cell tail broken --- reproduces across five simultaneous differences: chemistry (LFP $\rightarrow$ NMC-LCO), form factor (18650 $\rightarrow$ pouch), laboratory ($\rightarrow$ Oxford), load profile (constant current $\rightarrow$ Artemis dynamic discharge), and time grid (per-cycle $\rightarrow$ every-100-cycles characterization). The collapse is not an artifact of one dataset; with n = 8 we report full distributions and per-cell binomials and claim nothing percentile-shaped.

\textbf{The engine's silence is the second measurement of the ceiling.} On each LOO fold the calibration pool holds four cells; ESS\_batt saturates at 3.73 under the cap flag, the weights degenerate to cell balancing, and the engine's readings are digit-identical to plain conformal. Panel C's representativeness ceiling (§8.3) and this pool ceiling are the same sentence at two scales: \textbf{the repair capacity of hierarchical weighting is bounded by the diversity of batteries in the calibration pool, not by the number of cycles.} Fleets that want repairable per-cell promises should buy pool diversity, not cycle count.

\subsubsection{Reconciliation of preregistered predictions}\label{reconciliation-of-preregistered-predictions}

\begin{longtable}{@{} >{\raggedright\arraybackslash}p{(\linewidth - 4\tabcolsep) * \real{0.3333}} >{\raggedright\arraybackslash}p{(\linewidth - 4\tabcolsep) * \real{0.3333}} >{\raggedright\arraybackslash}p{(\linewidth - 4\tabcolsep) * \real{0.3333}}@{}}
\caption{P1--P10, written before execution, reconciled after.}\\
\toprule\noalign{}
\begin{minipage}[b]{\linewidth}\raggedright
\#
\end{minipage} & \begin{minipage}[b]{\linewidth}\raggedright
prediction (summary)
\end{minipage} & \begin{minipage}[b]{\linewidth}\raggedright
outcome
\end{minipage} \\
\midrule\noalign{}
\endhead
\bottomrule\noalign{}
\endlastfoot
P1 & Mondrian: batch tier repaired, cell tier untouched & $\checkmark$, sharpened: q10 repaired cosmetically while worst unmoved \\
P2 & GP: overconfident, worst-cell disaster & $\times$ reshaped: conservative via non-converged hyperparameters; conclusion (no trustworthy $\sigma$) holds with rewritten mechanism \\
P3 & BHM between base and Mondrian & half $\checkmark$: conservative coverage at 2$\times$ width; drift ``win'' is a 100\% fallback artifact \\
P4 & sim-only reverses the median & $\times$ reversed: no reversal under battery-level pools; cause relocated to pool narrowness \\
P5 & bal-only $\approx$ base, tail kept & $\checkmark$; plus unexpected +0.037 pooled on the drift panel \\
P6 & trajectory family hurts coverage most & reshaped to the width axis: 29$\times$ width, worst-cell collapse, q10 \emph{improves} \\
P7 & worst degrades monotonically in H & half $\checkmark$: two-axis reading (H=10 improves, H=50 degrades; width monotone 6$\times$) \\
P8 & comparator fallbacks fire on the 2018 batch & half $\checkmark$: Mondrian 100\% (digit-identical to base); BHM 100\% (different form: global scale) \\
P9 & cross-domain coverage collapses & $\checkmark$ (0.286) \\
P9b & engine does not save it & $\checkmark$ (0.376) \\
P10 & within-domain skeleton replicates & $\checkmark$ and exceeded (worst 0.686 at n = 8) \\
\end{longtable}

Three predictions reshaped under the data, each with a located mechanism; none left floating, none quietly rewritten. The two practice-machine readings that failed to replicate (H-sweep monotonicity, GP's failure form) were retired by this reconciliation rather than by a reviewer.

\subsection{Discussion and limitations}\label{discussion-and-limitations}

\textbf{The signature line: three tiers of promise, each with a measured anchor.} A ``90\%'' interval is not one promise; on nested data it is three, and they fail in different regimes. The \emph{guarantee} tier --- same station, slow drift --- is where the engine earns its keep: worst-cell coverage 0.642$\rightarrow$0.694 at the head configuration (0.014$\rightarrow$0.311 at the C optimum --- a diagnosis, not a selection; the head stays at 5) with the batch tier recovered to 0.884 and no deferrals. The \emph{assumption} tier --- a protocol change mid-fleet --- is where the frozen weights harm monotonically on the SOH scale (§8.3); we report the curve with no deferral behaviour to soften it and mark the escape route as operational, not statistical: a requalification window after which the new protocol's cells re-enter the calibration pool. The \emph{limitation} tier --- across a laboratory or chemistry boundary --- is empirical: 0.286 pooled coverage, no repair, no dressing up. A deployer should read the tier before reading the number; the same interval width is a guarantee in the first regime, a bet in the second, and a coin toss in the third.

\textbf{The repair ceiling is the pool, measured twice.} Panel C's cap census (43 of 83 test cells flagged at target 20, achieved ESS $\approx$ 18) and the Oxford folds (ESS\_batt saturating at 3.73 over four calibration cells) are the same sentence at two scales: the repair capacity of hierarchical weighting is bounded by the diversity of batteries in the calibration pool, not by the number of calibration cycles. The operational corollary is uncomfortable for data-collection practice: a fleet that wants repairable per-cell promises should spend its next budget on pool diversity --- more protocols, more chemistries, more form factors --- not on deeper cycling of the batteries it already has.

\textbf{Why we stop at panel B instead of chasing it.} Repairing drift \emph{inside} the freezing protocol would require weights that read residual scales --- exactly the data-dependent weighting that §5.5 forbids, and forbids on purpose: the auditability of the engine (weights computable from prediction-time-available quantities, nothing touching outcomes) is the property that makes it deployable at all. A weight that sneaks a look at residuals could likely buy back panel B, and would convert every future coverage number into an unauditable mixture of modeling and fitting. We name the trade-off and stop on the auditable side. The constraint is not an aesthetic choice: in regulatory and insurance contexts a coverage number whose weights cannot be recomputed from prediction-time quantities is not an admissible input --- auditability is a deployment precondition, not a preference. And the reported dead end is a finding, not a concession: repairing panel B inside the protocol is structurally impossible, a fact about the problem's shape rather than about our choices.

\textbf{RUL, one sentence.} The first cycle at which the SOH trajectory interval crosses the end-of-life threshold induces a remaining-useful-life interval at zero additional experimental cost; we performed no RUL-specific runs and claim nothing beyond the inducement.

\textbf{Limitations.} (1) \emph{Statistical power}: per-cell claims rest on exact binomial tests with Bonferroni correction over cells and on full-distribution reporting, but the test panels hold 25 cells per A seed, 40 on the drift panel, 83 on the official split, and 8 at Oxford --- the protocol bounds what any per-cell claim can carry, and percentile statistics on n = 8 are reported as distributions for that reason. (2) \emph{Same-chemistry cross-laboratory transfer} (e.g., NASA PCoE) is deferred: the Oxford arm crosses chemistry, laboratory, form factor, and load profile simultaneously; the isolated same-chemistry-different-lab cell of the transfer matrix remains unmeasured here. (3) \emph{One base model}: all formal runs use a ridge base learner; the engine is base-model-agnostic by construction, but we demonstrate exactly one base. (4) \emph{Comparator deferral asymmetry}: GP, BHM-lite, and Mondrian carry no deferral mechanism by design ($\infty$ $\equiv$ 0 in Table 3); the deferral column compares nothing for those arms, and we flag rather than impute it. (5) \emph{Unmapped envelope}: between the pool sizes where the engine degenerates to plain conformal (four calibration cells per Oxford fold) and where it separates from it (the Severson panels), the operating envelope is not charted --- we report the two endpoints, not the curve between them.

\textbf{The transferable artifact is the protocol, not the engine.} The three-tier signature line, the worst-cell binomial, and the deferral share are method-agnostic: they audit any interval producer, weighted or plain, Bayesian or conformal. The q10 falsification (§8.4 --- a statistic that can be ``repaired'' to 0.825 while the worst cell does not move) is the argument for making the worst-cell statistic mandatory. Wherever a per-cell interval promise is made --- warranty reserves, retirement scheduling, second-life grading --- the protocol travels with the numbers. One redesign is flagged for the record: the deferral trigger as configured is row-level and did not bind at these pool sizes; an ESS-keyed trigger is the natural next form, making deferral speak in the engine's own currency.

\subsection{Conclusion}\label{conclusion}

On nested battery degradation data, pooled conformal coverage conceals cell-level failure: fleet averages of 0.875--0.906 coexist with worst-cell coverage of 0.001--0.020, measured under a preregistered gate. Hierarchically weighted conformal prediction with frozen side information repairs part of that gap --- 10th-percentile cell coverage 0.750$\rightarrow$0.830 at zero deferrals --- and localizes the rest: protocol drift is a frozen-signature dead end on the normalized scale, and repair capacity is bounded twice over by calibration-pool battery diversity rather than cycle count. The boundary is empirical, not aspirational: cross-laboratory transfer collapses to 0.286, while the collapse skeleton replicates within-domain on a second chemistry (0.933 pooled, worst 0.686). What we leave behind is a reporting protocol --- three tiers of promise, worst-cell statistics, deferral share --- that makes per-cell interval claims auditable wherever they are made.

\begin{center}\rule{0.5\linewidth}{0.5pt}\end{center}

\subsection{Declarations}\label{declarations}

\textbf{CRediT authorship contribution statement.} Xiangyu Liu: Conceptualization, Methodology, Software, Validation, Formal analysis, Investigation, Resources, Data curation, Writing -- original draft, Writing -- review \& editing, Visualization, Supervision, Project administration, Funding acquisition.

\textbf{Funding.} This research did not receive any specific grant from funding agencies in the public, commercial, or not-for-profit sectors.

\textbf{Declaration of competing interest.} The author declares no known competing financial interests or personal relationships that could have appeared to influence the work reported in this paper.

\textbf{Data availability.} Both datasets are public and cited in the manuscript: the 124-cell LFP cohort \cite{severson2019} and the Oxford degradation dataset (\cite{birkl2017}; doi:10.5287/bodleian:ko2kdmygg). The analysis code, the preregistered protocols and gate criteria, the formal run logs, the figure-generation scripts and all derived coverage tables are archived at Zenodo: doi:10.5281/zenodo.23664797.

\textbf{Declaration of generative AI and AI-assisted technologies in the writing process.} During the preparation of this work the author used Claude (Anthropic) as an AI assistant for literature organization, for drafting and editing of text, and for development of analysis code. After using this tool, the author reviewed and edited the content as needed and takes full responsibility for the content of the published article. All figures were generated deterministically by Python scripts (matplotlib, fixed random seeds) archived with the code; no generative-AI image tools were used.

\begin{center}\rule{0.5\linewidth}{0.5pt}\end{center}

\end{document}